\documentclass[lettersize,journal]{IEEEtran}
\usepackage{amsmath,amsfonts}
\usepackage{algorithmic}
\usepackage{algorithm}
\usepackage{array}
\usepackage[caption=false,font=normalsize,labelfont=sf,textfont=sf]{subfig}
\usepackage{textcomp}
\usepackage{stfloats}
\usepackage[hyphenbreaks]{breakurl}
\usepackage[hyphens]{url}
\usepackage{verbatim}
\usepackage{graphicx}
\usepackage{cite}
\usepackage{float}
\floatstyle{plain}
\newfloat{mybox}{tbhp}{ext}
\floatname{mybox}{Box}
\begin{document}

\title{Is open robotics innovation a threat to international peace and security?}

\author{Ludovic Righetti, Vincent Boulanin
\thanks{This work was supported in part by a decision of the Council of the European Union on “Promoting responsible innovation in artificial intelligence (AI) for peace and security” (Council Decision (CFSP) 2022/2269).}
\thanks{Ludovic Righetti is with the Tandon School of Engineering at New York University, USA and the Artificial and Natural Intelligence Toulouse Institute, Toulouse, France (e-mail: ludovic.righetti@nyu.edu).}
\thanks{Vincent Boulanin is Programme Director at the Stockholm International Peace Research Institute, Stockholm, Sweden. (e-mail: boulanin@sipri.org).
}
}

\markboth{}%
{Righetti and Boulanin: Is open robotics innovation a threat to international peace and security?}


\maketitle

\begin{abstract}
Open access to publication, software and hardware is central to robotics: it lowers barriers to entry, supports reproducible science and accelerates reliable system development. However,  openness also exacerbates the inherent dual-use risks associated with research and innovation in robotics. It lowers barriers for states and non-state actors to develop and deploy robotics systems for military use and harmful purposes. Compared to other fields of engineering where dual-use risks are present – e.g., those that underlie the development of weapons of mass destruction (chemical, biological, radiological, and nuclear weapons) and even the field of AI, robotics offers no specific regulation and little guidance as to how research and innovation may be conducted and disseminated responsibly. While other fields can be used for guidance, robotics has its own needs and specificities which have to be taken into account. The robotics community should therefore work toward its own set of sector-specific guidance and possibly regulation. To that end, we propose a roadmap focusing on four practices: a) education in responsible robotics; b) incentivizing risk assessment; c) moderating the diffusion of high-risk material; and d) developing red lines. 
\end{abstract}


\section{Introduction}
\IEEEPARstart{T}{he} conflict in Ukraine has starkly illustrated the dual-use nature of robotics (see definition in Box 1). The ingenuity displayed by Ukrainian soldiers in repurposing and modifying commercially available drones for military applications has captured global attention \cite{NewYorkTimes2025}. The increasing affordability of commercial drones was certainly the main enabling factor, but it was not the only one. Advances in 3D printing and readily available open-source software and hardware have also made the production and modification of robotic systems easier and cheaper. Beyond Ukraine, this has notably increased the use of robotics by a much larger range of actors, including malicious actors such as rogue states, militias or terrorist groups.

Dual-use risks associated with robotics have been known for a long time and are regularly discussed in policy conversations on arms control and international peace and security (e.g. the intergovernmental process on Lethal Autonomous Weapons Systems at the United Nations Conventions on Certain Conventional Weapons\cite{laws}) and in debates within the technical and scientific community on responsible innovation and ethically aligned design\cite{ieee_globallyaligned}.

These conversations, however, have so far paid little attention to the role of openness and especially open-source software and hardware in the dual-use risk equation. Admittedly, the breakthrough of generative AI has sparked a vivid debate in the technical and scientific community, and in mainstream media, on the risk of harm associated with openly publishing certain learned models\cite{economist_aifree,llamachina}. That debate, however, has focused on advanced AI models and has primarily involved actors from the AI industry. To our knowledge, little has been written or said (at least publicly) about the dual-use challenges associated with openness in the specific context of robotics.

\begin{mybox}
    \begin{center}
    \fbox{\begin{minipage}{0.45\textwidth}
{\bf Box 1. Defining dual-use \cite{forge2010note}}\\
The term dual-use was coined in the 1990s by the US Office of Technology Assessment to underline that technologies that were underlying the development of weapons of mass destruction also had civilian and peaceful purposes. Over time, the use of the term has evolved. In arms control and export control circles, it commonly refers to technology that has both civilian and military use. In broader discussions about societal impact of technology, it is often used in an even broader sense to convey the general idea that technology may have ``an intended use or primary purpose which is good (or at least not bad) and a secondary purpose or use which is potentially harmful and is not intended by those who developed the technology in the first place''. This framing is intended to encompass a broader series of uses than military activity. We use that latter broad definitional approach here. 
\end{minipage}}
\end{center}
\end{mybox}

\begin{figure*}
    \centering
    \includegraphics[width=0.7\linewidth]{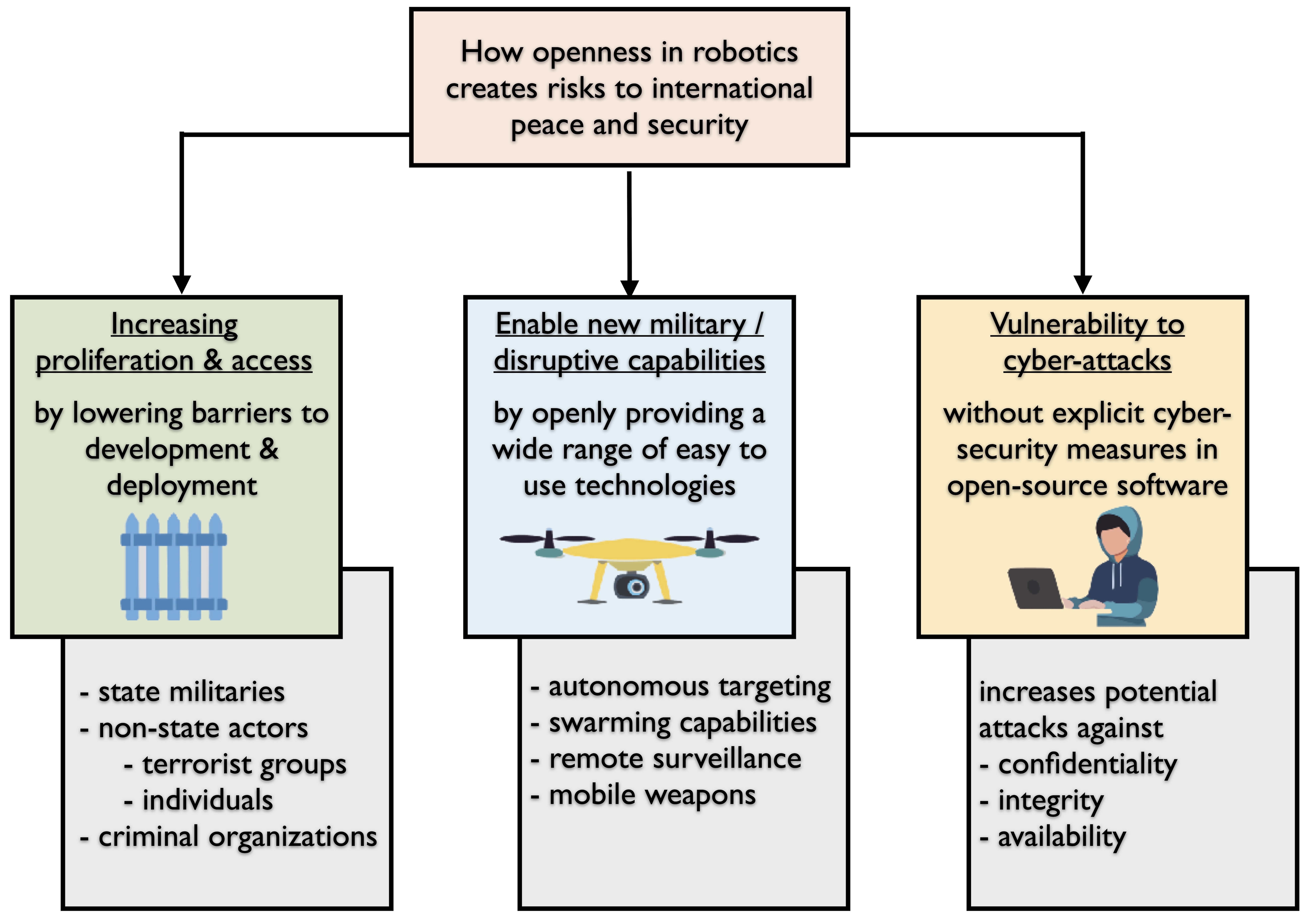}
    \caption{Risk pathways: how openness in robotics creates risk to international peace and security.}
    \label{fig:risks}
\end{figure*}
This article aims to lay the ground for an informed and nuanced discussion within the robotics community about the dual-use risks associated with open robotics. The immediate objective is to ensure that these risks are adequately considered in the conversations around  ``responsible robotics''. The longer-term objective is to support the emergence of norms and practices that can help roboticists uphold the principles of open research and innovation, which have been instrumental in the field's rapid progress, while effectively mitigating the inherent dual-use risks associated with robotics technology.

To that end, the article addresses three interrelated questions: 1) How openness in robotics may exacerbate dual-use risks associated with robotics research and innovation (Section \ref{sec:ii}), 2) What lessons can be learned from other disciplines as to how dual-use risk associated with openness may be mitigated (Section \ref{sec:iii}), and 3) What concrete steps could the robotics community take to ensure it is better prepared to address that problem (Section \ref{sec:iv}). 

\section{Openness and dual-use risks in robotics}\label{sec:ii}
Openness plays a pivotal role in accelerating the development of robotics technology. Open-access publication, software and educational content  democratize state-of-the-art knowledge, facilitate reproducibility, accelerate research, and, importantly, foster a collaborative international community of scientists.  There is no doubt that such resources should remain openly accessible.  However, openness can also exacerbate the dual-use risks and cyber-security risks associated with robotics and represents for that reason a challenge for arms control, proliferation and international peace and security more broadly (cf. Figure \ref{fig:risks} for an illustration of risk pathways).

\subsection{Openness already exacerbates dual-use and misuse}
Recent history already provides evidence that openness in robotics could lower barriers to entry for state and non-state actors to develop and deploy robotics systems for military use and harmful purposes, or be used to enhance certain capabilities in existing weapon systems such as vision-based navigation, autonomous targeting, or swarming\cite{Verbruggen_2019}.

ROS, an open software ecosystem originally intended for civilian purposes, now supports the development of military robotic autonomous systems through ROS-M
\cite{rosm}. Meta’s Llama models are reportedly used by both the US\cite{llamaus} and Chinese\cite{llamachina} military. Betaflight, an open source multi-rotor flight control software, is now widely used by Ukrainian drone makers to develop kamikaze drones\cite{Hambling_2024}.  For almost a decade now, non-state armed groups, like the Islamic State, have been misusing commercially available drones and open source software like QGround to create novel ways to wage war and terrorist attacks in the Middle East and counter no-fly-zones \cite{Veilleux-Lepage_Archambault_2022}\cite{balkan2017daesh}.

Obviously, military use does not necessarily equate misuse or harmful use. The point here is to recognize that the work roboticists make openly accessible can find totally different uses than the ones initially intended and that such uses can have implications for international peace and security.

\subsection{Openness and the cybersecurity risk}
Furthermore, unless proper security measures and development practices are adopted, open-source software for robotics systems may also be vulnerable to cyberattacks\cite{Botta_Rotbei_Zinno_Ventre_2023}. Hackers could exploit vulnerabilities to gain control of robots, causing them to malfunction or be used for malicious purposes. The Open Source Security Foundation and Open JS Foundation reported recently that unknown actors had attempted to insert a secret backdoor in the XZ Utils software distributed with Linux and urged all open-source software maintainers to take steps to protect their projects against such attacks\cite{nearmiss,openssf}. Furthermore, a majority of ROS and ROS-Industrial users recognize that they do not invest sufficiently in cybersecurity for their applications\cite{Mayoral-Vilches_2022}. In this case, unintended consequences stem rather from the lack of systematic cybersecurity measures by open-source developers rather than a dual-use problem.

\subsection{Openness, large robotics data and foundation models}
As robotics further integrates advances in machine learning, it will rely more on both large pre-trained models and datasets to train new models. While an open approach to datasets and models is critical for advancing research, it creates also additional cybersecurity and misuse risks \cite{jailbreak}. For example, open data repositories might be subject to  data poisoning. As open models are becoming more generalists with functionality across robot embodiments and sensing modalities, they also become increasingly  “misuse-ready”. Such models can provide various misuse possibilities as they require less engineering effort.

\subsection{Nuancing dual-use risks}
These risks should not eclipse the benefits of openness. In our view, unilaterally or indiscriminately closing research and innovation would be a very damaging response. Instead, we propose to look at risk mitigation strategies from other fields.

\section{Looking for guidance}\label{sec:iii}
\begin{figure*}
    \centering
    \includegraphics[trim={50 150 550 300},clip,width=0.65\linewidth]{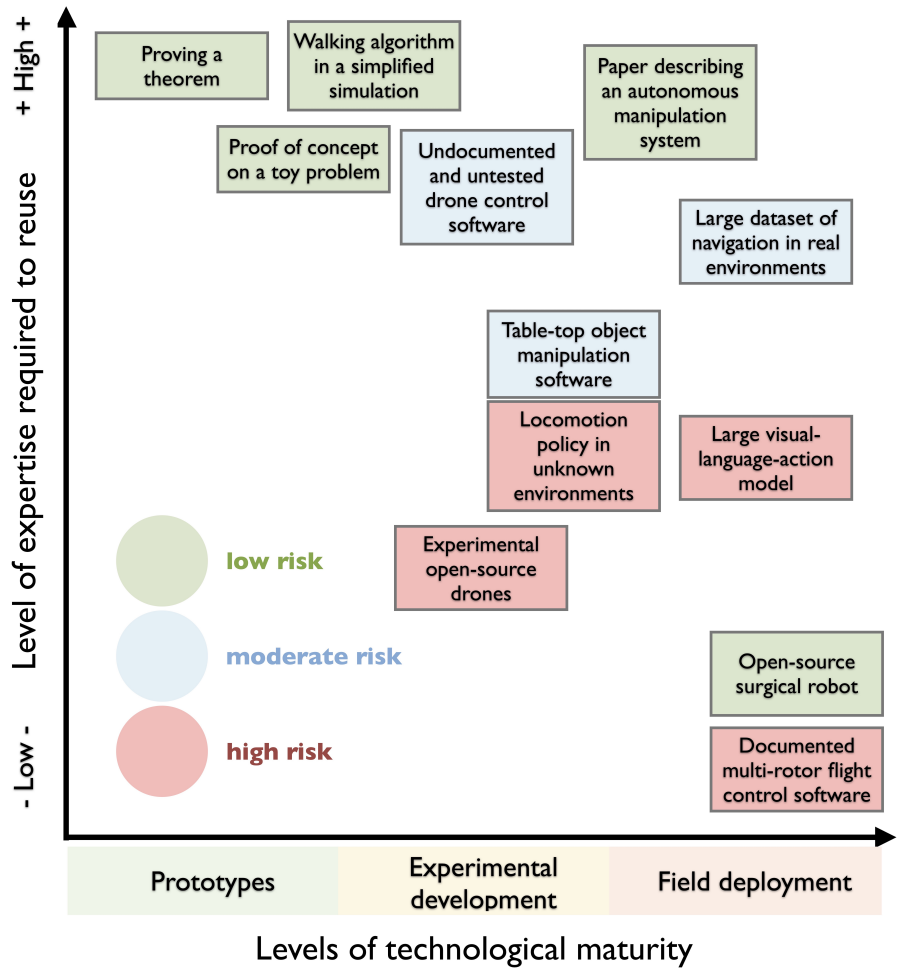}
    \caption{Illustration of how research and innovation in robotics present different levels of risk depending on technology maturity and the expertise needed to reuse the work for harmful purposes. While every example shown above could be misused in some way, the likelihood and potential scale of misuse needs to be factored into risk assessment. For example, an open-source surgical robot could be misused to cause harm but a drone flight control software poses higher risk as it is more likely to be directly misused and it can cause harm at a larger scale due to the provided mobility capabilities.}
    \label{fig:maturity}
\end{figure*}

\subsection{Lack of sector specific guidance }
Dual-use risks are a concern for many engineering fields. Some of them, especially for those that underlie the development of weapons of mass destruction (chemical, biological, radiological, and nuclear weapons), have developed clear guidance, and in some cases regulations, as to how research and innovation may be conducted and disseminated. Nuclear and biological research, for example, are strongly regulated, from the right to conduct certain types of research, where it can be conducted, whether it can be published, etc. These fields also have community-driven processes intended to mitigate dual-use risks associated with knowledge diffusion. For instance, BioRxiv and medRxiv, the preprint servers for biology and health sciences, screen preprint submissions for material that poses a biosecurity or health risk.

The field of robotics, in comparison, offers no specific regulation and little guidance as to how roboticists should address dual-use risks associated with open research and innovation. It is telling, for example, that there are no screening mechanisms for the submission of papers on arXiv. There are also no incentives for roboticists to do a risk assessment before uploading software on Github or presenting their findings in academic conferences. Dual-use risks are not even a topic that roboticists (or most engineers) are really invited to think of as part of their university education. This is problematic  for three reasons. 

First, contrary to nuclear and biological research, there is little that prevents malicious actors from misusing open robotics resources to cause harm. While the production of nuclear weapons and biological weapons is reserved only to a small number of resourceful and motivated actors (e.g. states), the production of “do-it-yourself” mobile weapon systems using open-source design and software and off-the-shelf commercial components is within the reach of any motivated individual or group.

Second,  roboticists are bound to face dual-use concerns at some stage in their careers. For instance, they will likely be confronted with situations where they would have to assess whether their work is subject to export control regulations on dual-use items\cite{Boulanin_Brockmann_Richards_2020}. Some funding organizations may also require them to conduct an impact assessment that include mapping out possible downstream consequences\cite{eu-ethics-assessment}. 

Third, guidance and practices stemming from other fields of sciences and engineering (e.g. on generative AI\cite{pai}) may be informative but not all are directly transferable to robotics. Robotics has its own specificities (i.e. the combination of physical moving objects with autonomous decision-making capabilities), including with regards to the difficult trade-off between benefits and risks of open research and innovation.

We therefore argue that the robotics community should work toward its own set of sector specific guidance and possibly regulation (cf. Section \ref{sec:iv}).
However, roboticists who desire to engage on the topic urgently, for instance because they work on projects with great dual-use potential, should not feel that they need to wait from sector-specific guidance to take action. A lot can be learned from the literature on risk management and from how other disciplines mitigate dual-use risk stemming from openness \cite{rausand2013risk}. 

Risk management typically involves three steps: risk identification and analysis, risk evaluation, risk mitigation. Each step involves answering a series of standard questions using methods appropriate to the problem (e.g. scenario analysis, fishbone methods, delphi technique, risk matrices). We outline below how these questions could be applied to the specific case of dual-use risks stemming from open robotics. 

\subsection{Risk identification and analysis}
Risk identification and analysis involve answers to three basic questions: a) what could go wrong; b) what is the likelihood of that happening; c) what are the consequences.

Identifying how open robotics research and innovation could have unintended negative consequences is not necessarily difficult as the general dual-use risk scenarios are known. Some actors may misuse open robotics for various harmful purposes: military operation, terrorist attacks, criminal activity or surveillance. Determining the likelihood of misuse and associated harm is less straightforward as it requires consideration of multiple factors, such as:
\begin{itemize}
    \item \emph{The nature and maturity of the work:} Some areas of robotics are more likely to be misused than others. Research on swarming presents more opportunity for malicious actors than research on surgical robots. Similarly, a theoretical research paper is less ‘misuse ready’ than a mature software package, as it would require more efforts to misuse it.
    \item \emph{Dissemination form:} The likeliness of misuse depends also on  how the work is disseminated and to whom. Whether the work is disclosed via an open-source repository, pre-compiled library with limited API,  paper in beyond-paywall journals matter as it determines by who and how easily the work can be accessed.  The audience and format also count. Work intended for the research community, developers’ community and public consumption require different levels of technical knowledge and resources for potential misuse. Moreover, the fact that some work is openly accessible does not mean that it can be easily misused. Without proper testing and documentation, data and algorithms may be difficult to reuse even for people with technical expertise. 
\end{itemize}
These factors ultimately boil down to the question of what resources and methods will be needed to access, modify and misuse the work. Figure \ref{fig:maturity} provides an illustration of these relationships. 

Foreseeing potential harm is the other critical variable in the equation. Factors to be considered include: 
\begin{itemize}
    \item \emph{What or who is impacted} by the possible misuses – ranging from values, objects, processes to people. 
\item \emph{The temporal and geographical dimension of the impact} – e.g. whether it will be a one-off localized event or something that will happen repeatedly and have large scale effects.  
\item \emph{The order of effect} – whether harm is directly caused by the misuse or whether harm may result from second or third order effects from that misuse. 
\end{itemize}
Sometimes, identifying and analyzing risk is relatively easy. For instance, it is not hard to foresee that an open-source software for vision-based navigation could be used for drone surveillance and drone strikes. In some other cases –– notably when it comes to research publications on more fundamental or abstract problems – it may be harder to identify possible misuse and associated harm as downstream applications may be multiple and in some cases not be clear yet. It is also a process that may require domain knowledge. This is one of the reasons why the literature on risk assessment typically recommends making the process participatory\cite{grunwald2018technology}.

\subsection{Risk evaluation}
\begin{figure}
    \centering
    \includegraphics[width=0.9\linewidth]{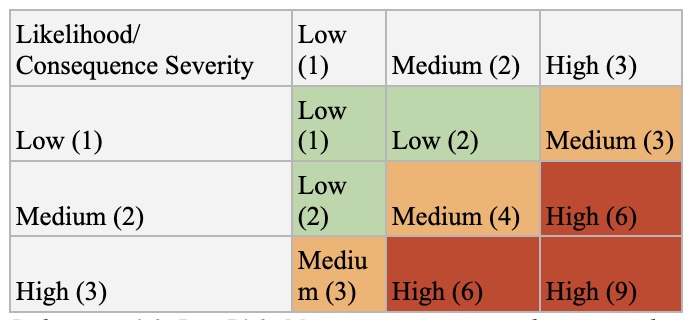}
    \caption{Example of a 3x3 Risk Matrix. Risk rating: {\bf 1-3: Low Risk:} May require no action or basic control measures; {\bf 4-6: Medium Risk:} Requires implementation of risk reduction measures; {\bf 6-9: High Risk:} Requires action to drastically reduce or avoid risk altogether.}
    \label{fig:eval}
\end{figure}

Risk evaluation assesses the degree of severity or tolerability and determines what can be done about it. Risk matrices are commonly used for that purpose but there are other tools available (c.f. Figure \ref{fig:eval}).

This typically leads to four options:
\begin{itemize}
    \item[1] \emph{Accept the risk.} No particular measures are needed before releasing the work.
\item[2] \emph{Avoid the risk.}  The possibility of massive misuse and severe harm may require drastic measures, including  pausing or not publishing the work. 
\item[3] \emph{Mitigate the risk.} Medium and high level risks typically involve adopting measures that will reduce the likelihood of misuse or limit the scale of harm. 
\item[4] \emph{Transfer the risk.} People behind the work may also decide to transfer risks horizontally (e.g. raising awareness in their community) or vertically by alerting governments or other decision makers. 
\end{itemize}
The risk stemming from openness also needs to be weighed against possible benefits such as scientific progress, impact on the research and innovation ecosystems (e.g. extent to which it will democratize and speed up innovation) and benefits to society (e.g. improvement in health, safety, environmental sustainability). 

The risk/benefit equation is challenging because it necessarily involves a part of subjectivity and may be perceived differently from one actor to the other.  Commercial actors may have different interests and priorities than academics. End-users or beneficiary of the technology  are also likely to have different views than the people that may be impacted by its misuse. Here again the literature recommends that this evaluation process be deliberative and include a diversity of views\cite{schiff2020principles}. Transparency is also valuable so third parties can understand what motivated the conclusions. 

\subsection{Risk mitigation}
Risk mitigation measures are typically tailored to the nature of the work and risk. A lot can be learned from how other fields of science and technology, not least the field of biological research, have tackled dual-use risks stemming from open research and innovation\cite{resnik2006openness,tucker2012innovation}. We highlight here two general lessons. 

First, openness can be seen as a continuum rather than a binary opposition between open and closed research and innovation. There are a number of potential options to limit risk without hampering the benefit of openness:
\begin{itemize}
    \item \emph{Curate the information} to publish carefully and remove problematic details (e.g. the approach taken in \cite{urbina2022dual} when discussing the dual-use risk associated with AI powered drug discovery),
\item \emph{Differential privacy} to protect sensitive data,
\item \emph{Staged release} – when components of the work are gradually released to allow time to monitor for misuse and implement safeguards. That’s the approach that OpenAI followed when it deployed GPT2 to GPT4,
\item \emph{Controlled access} - follow the know-your-customer principle by providing access only to trusted researchers or organizations that agree with specific terms of use,
\item \emph{Limit functionality} - release of a version of the work with reduced capabilities to limit potential misuse,
\item \emph{Deploy technical safeguards} including geo-fencing, monitoring and detection systems and anti-tamper mechanisms, a common practice in industries that produce life-critical systems, not least weapon systems \cite{Defense_Security_Cooperation_Agency},
\item \emph{Licensing and code of conduct} to prohibit certain uses or require adherence to specific ethical guidelines. This is already a standard practice among major AI and robotics companies.  Boston Dynamics, for instance, prohibit the weaponization of its general purpose robots\cite{Boston_Dynamics_ToR}, 
\item \emph{Risk transfer through advocacy} - engage with policy makers to advocate for regulations or actions that may reduce the likelihood of misuse or reduce the level of harm. 
\end{itemize}

The second lesson is that the implementation of such measures is dependent on the existence of an organizational culture and processes that support the conduct of risk management in universities, research institutes and companies. Researchers and engineers need to be sensitized to the risks, encouraged to actively think of the risks associated with their work, made aware that they do not have to choose merely between fully open vs fully closed research. That requires the mobilization of resources on the part of universities and companies for training but also for the development of institutional processes for risk assessments. Major universities and companies already do have teams and processes dedicated to this purpose, in the form of ethical review boards and compliance teams. The EU Commission has also produced a guidance document on how to complete an ethics self-assessment which covers AI broadly as well as the problem of weaponization and misuse of technology in general\cite{eu-ethics-assessment}. Such support infrastructures and guidance documents, however, are generally very high-level, seldom address robotics needs specifically and seem to remain relatively unknown within the robotics community. Therefore, we see value for the robotics community to work on guidance material that is tailored to its own needs.

\section{A roadmap}\label{sec:iv}
\begin{figure*}
    \centering
    \includegraphics[width=0.7\linewidth]{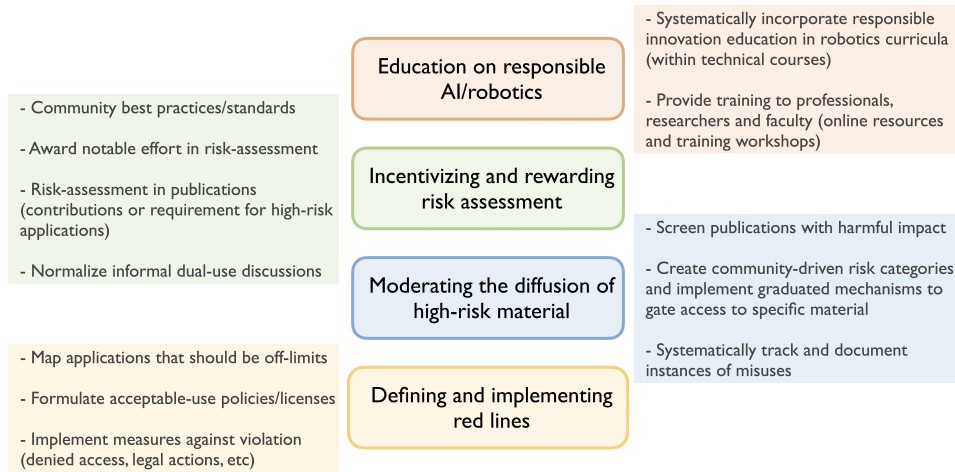}
    \caption{Overview of the proposed roadmap for risk assessment and mitigation. Four types of practices are identified and a summary of associated concrete actions are listed for each of them.}
    \label{fig:roadmap}
\end{figure*}

We now propose a roadmap with four avenues of action to help reduce the dual-use risks while fostering openness (Figure \ref{fig:roadmap}).
\subsection{Education on responsible AI/robotics}
Education is a requisite to enable other actions. Most technical universities do not provide dedicated classes on responsible research and innovation nor make students aware of dual-use risks of this type of technology\cite{Dignum_2021,Taebi_2019}. As a result, many roboticists are ill-equipped to meaningfully understand potential consequences of their work. Furthermore, classes on ethics or societal consequences of technology are often separated in the curriculum and can be perceived as unrelated to engineering practice. 

We suggest systematically integrating courses on responsible research and innovation in robotics education and, more importantly, to also include material on dual-use risk and cybersecurity within core robotics courses, e.g. while discussing the design of a robotic system for a specific application. To enable this, it is necessary to first educate current robotics teachers, researchers and professionals. It can be done by diffusing open-source material on responsible innovation \cite{unoda, RAI-NYU} and by organizing training workshops in robotics conferences and universities. Rendering responsible research and innovation education an integral part of robotics curricula would help engineers integrate these considerations in their practice, make informed decisions in their work and research directions \cite{Figaredo_Stoyanovich_2023} and unlock their ability to be proactive in this discussion. These recommendations should be rather straightforward to implement by individuals and universities without specific national or international coordination.

\subsection{Incentivizing and rewarding risk assessment}
The second avenue of actions consists in incentivizing and rewarding risk assessment as a critical component of responsible publication of robotics research and responsible deployment of robotics innovations. Individuals and organizations should be encouraged to assess the downstream negative consequences of making their work fully or partially openly accessible. They could be encouraged to do so both in formal and less formal ways.

Funding agencies could, for instance, demand such assessment as a formal condition for project funding. Professional organizations, like the IEEE Robotics and Automation Society (RAS) or the International Federation of Robotics, could also adopt and promote best practices in risk assessment as standards (e.g., through the IEEE RAS Standing Committee for Standards Activities), but also provide tools and decision frameworks that roboticists can rely on to identify, assess, and mitigate possible risks. Such tools could take the form of self-assessment checklists for individual researchers, guidance as to how faculties and labs can set up ethical review boards. Editors of academic journals and organizers of major conferences could also play an important role. They could ensure that serious risk-assessment, where relevant, is valued as a significant contribution for publication and is peer-reviewed accordingly. It could also become a required part of papers discussing high-risk applications.  Setting standards or mandatory risk-assessment are institutional activities that may require a difficult to reach consensus. On the other hand, dedicated awards could officially recognize important contributions towards risk assessment and mitigation with very low institutional overhead.

Risk assessment can also be incentivized and rewarded in more informal ways. People in leadership positions – such as PhD supervisors and heads of laboratories – can for instance build ad hoc opportunities for students and researchers to think of possible risks. They can convene seminar discussions on the topic, provide introductions to external experts and stakeholders (e.g. social scientists, cyber-security experts), ask specific questions on this topic as part of their mentoring. These activities can be initiated in a bottom-up manner by individuals interested these issues.

\subsection{Moderating the diffusion of high-risk material}
Further, the robotics community could self-regulate its openness using mechanisms inspired from other sectors. One could think of screening publications before they appear on arXiv and in robotics conferences and journals to prevent the publication of content posing serious risks of harm. Similarly, one could think of graduated mechanisms that “gate” access to certain source-code or data on open sources-repositories\cite{Solaiman_2023}. One example of that would be to require people to identify and authenticate themselves via a trusted source and indicate the intended end-use. This is already  common practice in the gene synthesis industry. Accessible tools, such as open applications to provide a gating mechanism for certain, at risk, repositories would help facilitate the deployment of these strategies.

The moderation of publications and gating of high-risk open-source repositories is certainly a sensitive topic. Clear rules would need to be established to ensure that the free pursuit of scientific knowledge is not impacted and that labs, companies or governments would not instrumentalize such processes to close their research. This would necessarily need to be a community effort. Organizations such as RAS and euRobotics could for example  come up with their own categories of risk levels. AI companies have made an effort over the past years to develop AI risk levels following the model of biosafety levels in life science. The robotics community could engage in a similar exercise and rank robotics research and applications that may be deemed low risk, medium and high risks or simply unacceptable. Each category would then be coupled to different risk prevention and mitigation measures. Under such a scheme, low and medium risk categories would be subject respectively to no and partial requirement in terms of dual-risk screening and gating, while the high-risk category may demand additional risk mitigation measures and applications in the unacceptable category would not have access to mainstream venues. 

Lastly, we suggest to systematically track and document real cases of dual-use and misuse of robotics in order to understand the scale of the risks and potential patterns of usage (e.g. which technology, where, for which application) in order to help in the creation of meaningful mitigation strategies. This monitoring activity could for example be handled by groups such as the IEEE RAS Responsible Research and Innovation in Robotics and Automation Committee. 

\subsection{Defining and implementing red lines}
The robotics community could also seek to define and enforce its own red lines for the development and deployment of robotics technologies. Efforts to self-regulate and define red lines have already been made in that direction, notably in the context of the IEEE global initiative on ethics of autonomous and intelligent systems\cite{ieee_globallyaligned}. Companies such as Boston Dynamics, Unitree, Agility Robotics, Clearpath Robotics, ANYbotics and Open Robotics have called in an open letter for a regulation on the weaponization of general purpose robots\cite{bdi}. Such efforts were, however, very narrow in scope and there is certainly value for the community to further map end-uses of robotics that should be deemed off-limits or demand extra caution. It is beyond dispute that it will be a difficult endeavor for the community to agree on common red lines, not least because what is considered ethically acceptable or problematic is highly subjective. For comparison, States have been discussing how to govern the development and use autonomous weapons systems at the United Nations since 2014, and they are still widely divided as to what aspects of the development and use of such systems should be regarded as off-limits\cite{SIPRI_Essay}.  These differences of views are not insurmountable, however. States were, for example, able to agrees on a series of  high level principles on autonomous weapons back in 2019\cite{CCWGuidingprinciple}. The robotics community can make progress toward the identification of red lines incrementally. To support that process individuals and companies can engage in a reflection at their level with regard to what they consider to be non-acceptable use of their work. This could result in the formulation of acceptable use policies or terms of use that beneficiaries of open research and open-source design software would have to formally agree to (e.g., in the form of specific open-source licenses). This would provide a basis for revoking access, denying software updates, and potentially suing or blacklisting people who misuse the technology. Such measures are already implemented, to some extent, by some industry actors, not least Boston Dynamics. They could also be replicated by individuals and research organizations conducting open research.  Such type of red lines would remain ``self-regulatory'' and therefore might not provide a sufficiently robust legal basis to make actors that breach them accountable. The upshot is that these red lines could be implemented globally without being limited to specific national jurisdictions.     

\section{Moving forward}\label{sec:v}
The robotics community needs to take greater ownership of the discussion on the peace and security risks associated with open research and innovation. It is important to ensure that robotics research benefits society globally and does not become a driver of instability in the world. This goal, we believe, aligns with a majority of our community, including RAS’ mission to ``benefit humanity''. 
The robotics community also needs to be proactive on these issues for its own sake, to prevent any backlash from society and to preempt potentially counterproductive state or international regulations that could harm open science.  

This article aims to serve as an entry point for roboticists interested in the topic. Importantly, our roadmap does not advocate for increased regulation or administrative burden as much as it calls for a fundamental cultural shift in research practices.
It purposely offers recommendations that can be implemented by all, from individuals to professional societies. We hope it can serve as a productive tool to ground discussions to strike a proper balance between open science and innovation and dual-use risk mitigation. 

\section*{Acknowledgments}
The authors would like to thank C. Ovink, N. Mansard and O. Stasse for their comments on an earlier version of this manuscript.

\bibliographystyle{IEEEtran}
\bibliography{biblio}
%


 





\end{document}